\documentclass[preprint]{acm_proc_article-sp}

\usepackage{algorithm}
\usepackage{algpseudocode}
\usepackage{caption}

\usepackage{color}

\begin{document}


\title{A logical model of Theory of Mind for virtual agents in the context of job interview simulation}



%
%
%
%

%

\numberofauthors{2}

\author{
\alignauthor
Marwen Belkaid \\
       \affaddr{ETIS, UMR 8051}\\
       \affaddr{95000 Cergy Cedex, France}\\
       \email{marwen.belkaid@ensea.fr}
\alignauthor
Nicolas Sabouret \\
       \affaddr{LIMSI, CNRS, UPR 3251}\\
       \affaddr{91403 Orsay Cedex, France}\\
       \email{nicolas.sabouret@limsi.fr}
}

\maketitle

\begin{abstract}
Job interview simulation with a virtual agents aims at improving
people's social skills and supporting professional inclusion. In such
simulators, the virtual agent must be capable of representing and
reasoning about the user's mental state based on social cues that
inform the system about his/her affects and social attitude. In this
paper, we propose a formal model of Theory of Mind (ToM) for virtual
agent in the context of human-agent interaction that focuses on the
affective dimension. It relies on a hybrid ToM that combines the two
major paradigms of the domain. Our framework is based on modal logic
and inference rules about the mental states, emotions and social
relations of both actors. Finally, we present preliminary results
regarding the impact of such a model on natural interaction in the
context of job interviews simulation.
\end{abstract}







\section{Introduction and positioning}
During the last decade, several projects have proposed to use
intelligent virtual agents in digital games for user empowerment
\cite{Marsella2003, Paiva2004, Pareto2009, tardis2013,
  Batrinca2013}. The work presented in this paper considers the use of
virtual agents in job interview simulation games for young unemployed
peoples, a.k.a NEETs\footnote{NEET is a government acronym for young
  people not in employment, education or training. According to
  Eurostat, in march 2012, 5.5 million of European youngster (16 to 25
  years old) were unemployed meaning that $22.6\%$ of the youngster
  global population in European union is unemployed. This unemployment
  percentage is 10 points superior to the whole population showing
  that the employment of NEETs is a real problem in Europe.}. Current
research reveals that NEETs often lack self-confidence and the
essential social skills needed to seek and secure employment
\cite{Bynner2002}. Training with a virtual agent can help them acquire
self-confidence and improve their social skills. Indeed, it has
already been proven that training at job interviews with a virtual
agent could improve the performance \cite{hoque2013}.

The role of the virtual agent in such training games is to be able to
react in a coherent manner: based on the non-verbal inputs (smiles,
emotion expressions, body movements), the agent must select relevant
verbal and non-verbal responses. In this context, several work
illustrated the role of emotion regulation in the context of job
interviews. For instance, in \cite{Sieverding2009}, a study shows that
people who tried to suppress or hide negative emotions during a job
interview are considered more competent by evaluators. Similarly,
Tiedens \cite{Tiedens2001} shows that anger and sadness play an
important role in job interviews. For this reason, credible
simulation of emotions appears as a key issue when it comes to using
virtual agents in job interview simulations.

Most existing models for virtual agents rely on a reactive approach,
in which the system does not manipulate or reason on the mental states
of the interlocutor \cite{hoque2013,Paiva2004,Pareto2009}. However, in human psychology, \textit{Theory of
  Mind} (ToM) refers to the ability of human beings and primates to
interpret, predict and even influence others' behavior
\cite{baron1997mindblindness}. Such an ability is a key feature in the
development of intelligent virtual agents in the context of tutoring
and training systems. In this paper, we propose a new model of ToM for
virtual agent in the context of job interview simulation.

The next section briefly discusses existing research that serves as a
basis to our work. Sections \ref{sect:archi} and
\ref{sect:logic} present the general architecture and the logical
framework for our ToM. Section \ref{sect:implem} describes our implementation of
this model in the context of job interviews simulation. An outline of
the preliminary evaluation we conducted is given in Section
\ref{sect:eval}. Finally, results and perspectives are discussed in
Section \ref{sect:discuss}.
\section{Related work}

In order to be able to reason on the affective dimension of the
interaction, conveyed by the non-verbal behaviour of both
interlocutors, several models rely on the cognitive structure of
emotions and appraisal theories such as CPM \cite{scherer2010emotion}
or OCC \cite{Ortony1990}. These theories provide domain-independant
descriptions of triggering conditions of emotions, that are required
for the development the affective aspect of the ToM reasoner. For
instance, \cite{Adam2009,Dastani2012} are BDI-based implementation of
the OCC theory. FAtiMA's double appraisal model \cite{aylett2008if},
although not implemented using a BDI framework, also encodes the OCC
model. However, in these models, the inference mechanism itself
encodes the chosen Appraisal Theory. On the contrary, in our model, we
propose a theory-independant ToM reasoner. While our experiments were
conducted using an OCC-based model, the corresponding rules (described
in equation \ref{eq:emo}) could be easily replaced by another theory.

To support such adaptability, we propose to rely on the BDI model.
Several computational models of emotions have already been proposed
(e.g. \cite{herzig2002logic,Dastani2012}) that show that
BDI is a good basis to represent and to reason about the
interlocutor's mental state. Our aim is thus to define a logical model
of emotions and ToM in BDI.

From the philosophical point of view, a debate about
how ToM is processed by human adults opposes two theories. The
\textit{theory-theory}(TT) argues for a folk-psychology reasoning,
i.e. a set of rules one acquires regarding human mind
functioning. \cite{botterill1999philosophy}. The
\textit{simulation-theory} (ST) \cite{goldman2006simulating} defends a
mirroring or projection process allowing for taking someone else's
perspective. Various research demonstrated that neither pure TT nor
pure ST were realistic \cite{vogeley2001mind} and both theorists and
simulationists turn toward more hybrid models \cite{botterill1999philosophy}\cite{goldman2006simulating}.

Existing computational ToM models either imply a choice between the TT
and ST theories (\textit{e.g.} \cite{aylett2008if} that relies on a ST
approach, or \cite{bosse2011recursive} and \cite{pynadath2013you} that
position in the TT) or implement them separately as in
\cite{harbers2011explaining}. In our work, we propose a hybrid
approach that relies on theory-theory to model the agent's mental
states and commonsense rules, but also on simulation-theory to others'
perspective by projecting attributed mental states on its own
inference engine. Both models are integrated in the same reasoner.

\section{Architecture Overview}
\label{sect:archi}

\begin{figure}[b]
\centering
\includegraphics[width=\linewidth]{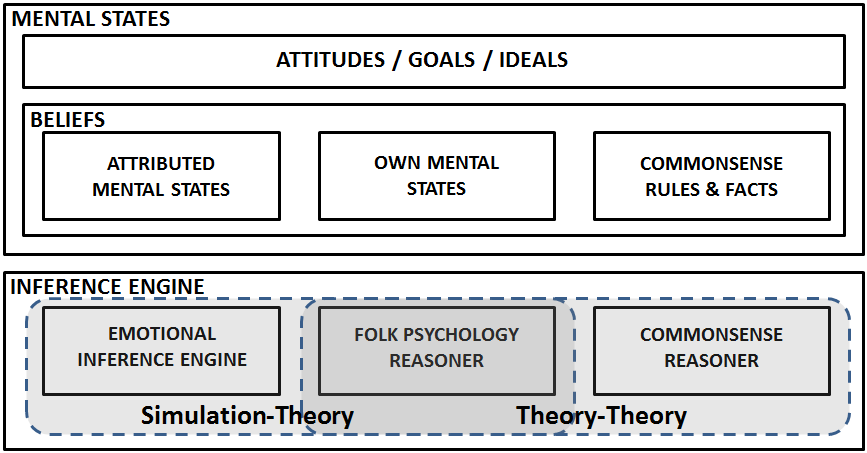}
\caption{General architecture allowing for modeling hybrid Theory of Mind. }
\label{fig:archi}
\end{figure}
Our ToM reasoning architecture consists of two main components, as presented on Figure \ref{fig:archi}.

The \textbf{agent's mental states} contains beliefs, attitudes, goals and intentions. Beliefs represent knowledge about general facts, rules of the world (\textit{i.e.} commonsense knowledge from the theory-theory) and mental states of self or other's (\textit{i.e.} attributed mental states). Attitudes represent appreciations of the current state of affairs and, by extension, desires (what the agent wants to be true in the future) and ideals (what the agents would like to be always true). Goals and intentions form the deliberative aspect, limited to immediate actions.

The \textbf{agent's inference engine} contains three parts. The folk-psychology deliberative reasoner is responsible for intention generation (according to the agent's beliefs and attitudes) and updating mental state. The Commonsense reasoner's role it to enrich the agent's beliefs base using commonsense rules and facts. Finally, the emotional inference engine computes emotion based on the appraisal theory.

Our hybrid ToM modeling relies on: 1) a TT approach based on folk-psychology and commonsense to reason about others, and 2) a ST approach consisting in projecting their attributed mental states on the agent's own inference engine. The following section details this logical model.

\section{Logical framework}
\label{sect:logic}

In the following, $\stackrel{\mathrm{def}}{=}$ and
$\stackrel{\mathrm{def}}{\Longrightarrow}$ respectively mean
\textit{equals by definition} and \textit{implies by definition}. The
former is used to define new operators as functions of others and the
latter to express inference rules.


\subsection{Syntax}
Assume finite sets of atomic propositions $ATM$, physical actions
$ACT$, illocutionary (speech) acts $ILL$, agents $AGT $, emotions
$EMO$ (which is a subset of the twenty two OCC emotions in our model),
and the intervals of real numbers $DEG = [-1,1]$ and $DEG^+=
[0,1]$. $ATM$ describes facts or assertions (\textit{e.g.}
\textit{salary\_is\_bad}, \textit{picnic\_is\_fun}) or external events
such as \textit{rain\_starts\_falling}. $ACT$ describes actions that
the agents or humans ($AGT$) may perform,
e.g. \textit{introduce\_itself} or \textit{have\_a\_picnic}.

Our model defines events as acts in which at least one of the actors
of the interaction take part. Elements in $EVT$ are tuples in
$AGT\times AGT\times (ACT\cup ILL(ATM))$ where the first element is
the actor that performs the action, the second is a passive agent and
the act can be either an actions ($ACT$) or a speech act ($ILL$). This
representation is similar to the one in \cite{ochs2009simulation}
except we associate a subjective \textit{degree of plausibility} as is
usually done in BDI models and we do not distinguish actions from
communcation. Illocutionnary speech acts have the form
$\varsigma(\varphi)$ and mean \textit{\textquotedblleft actor utters
  $\varphi$ to recipient through the illocutionary act $\varsigma$"}.

The language we define is the set of formulas described by the
following BNF (Backus-Naur-Form):
\begin{equation}
\begin{split}																				\\
Evt 	&: \epsilon	::= \langle a, (a|\varnothing), \alpha \rangle 	\,|\,	\langle a, a, Spk(\varsigma ,\varphi) \rangle	\\
Prp	&: \pi	::=	 p \,|\, \epsilon \,|\, Like_{a,b}^k \,|\, Dom_{a,b}^k 							\\
Fml		&: \varphi ::=  \pi \,|\, Bel_a^l(\varphi) \,|\, Att_a^k(\varphi) \,|\, Int_a(\varphi) \,|\, Emo_{a,(b|\varnothing)}^i(\varepsilon,\varphi) \,| \\
&\,\,\,\,\,\,\,\,\,\,\,\,\,\,\,\,\,\,\,\, N(\varphi) \,|\, U(\varphi,\varphi) \,|\, \neg\varphi \,|\, \varphi \wedge \varphi			
\end{split}
\label{eq:bnf}
\end{equation}
where $a,b \in AGT$, $\alpha \in ACT$, $p \in ATM$, $\epsilon \in
EVT$, $\varepsilon\in EMO$, $\varsigma\in ILL$, $l,i \in DEG^+$, $k
\in DEG$. $Like$, $Dom$, $Bel$, $Att$ and $Int$ are modal operators
and $N$, and $U$ are temporal operators \textit{Next} and \text{Until}
from LTL and CTL* \cite{pnueli1977temporal}. The other temporal operators $F$ and $G$ and
boolean conditions $\top$, $\bot$, $\vee$ and $\Rightarrow$ are
defined in the standard way. Moreover, in the events' representation,
we use ``$-$'' as the \textit{any} operator.

For the representation of social relation \cite{leary1957interpersonal}, $Like_{a,b}^k$
determines the level of liking agent $a$ has for agent $b$, while
$Dom_{a,b}^k$ represents the degree of dominance.

$Bel_a^l(\varphi)$ is a graded belief, in a similar manner to \cite{Dastani2012}, and has to be read \textit{\textquotedblleft a believes that $\varphi$ with certainty l"}. For instance, $Bel_a^{1}(\varphi)$ means \textit{\textquotedblleft a is sure that $\varphi$"} and $Bel_a^0(\varphi)$ can be read \textit{\textquotedblleft For a, $\varphi$ is not plausible at all"}.

Similarly, $Att_a^k(\varphi)$ is a graded attitude that has to be read
\textit{\textquotedblleft a appreciates/values the fact that $\varphi$
  with a degree l"}. In our context, this operator will be used to
cover various notions, such as \textit{Desires}, \textit{Ideals} and
\textit{Goals} that are represented with distinct modal operators in
other work such as \cite{Adam2009} and \cite{Guiraud2011}. We define
desires as a positive attitude toward future facts and ideals as what
the agents would like to be always true:
\begin{equation}
\begin{split}
Des_a^k(\varphi) &\stackrel{\mathrm{def}}{=} Att_a^k(F(\varphi)) \\
Ideal^{k>0}_a(\varphi) &\stackrel{\mathrm{def}}{=} Att^{k>0}_a(G(\varphi)) = Des^{-k<0}_a(\neg \varphi)
\label{eq:desideal}
\end{split}
\end{equation}
The definition of goals through attitudes will be presented in the
next subsection. 

Note that in our model, the subject of an attitude can as well be
\textit{preserving\_forest}, \textit{being\_nice\_to\_others},
\textit{hiring\_new\_employee} or $Bel_b^l(\langle a, c,
give\_sandwich\rangle)$, eventually encapsulated in temporal
operators.

As in classical BDI, $Int_a(\varphi)$ represents an agent's plan \cite{rao1991modeling} and has to be read \textit{"a intends to make $\varphi$ true"} (with $\varphi$ being an event in the general case).

$Emo_{a,(b|\varnothing)}^i(\varepsilon,\varphi)$ represent emotions. Following classical literature \cite{frijda1986emotions}, our emotions are related to facts and can be directed toward an agent. $Emo_{a,(b|\varnothing)}^i(\varepsilon,\varphi)$ has to be read \textit{\textquotedblleft a feels $\varepsilon$, eventually for/towards b, with intensity i, regarding the fact that $\varphi$"} with $\varepsilon\in EMO$. In the following sections, it will be simplified into $\varepsilon_{a,(b|\varnothing)}^i(\varphi)$.

For the sake of readability, we introduce new operators to represent agents' involvement in an event. $Resp_a$ expresses a \textit{direct responsibility}. Unlike \cite{Adam2009,Guiraud2011}, we do not consider an agent responsible for a situation it could have avoided. $Wit_a$ means that the agent witnessed the occurrence of the event:
\begin{equation}
\begin{split}
Resp_a(\epsilon) &\stackrel{\mathrm{def}}{=} (\epsilon = \langle a,-,-\rangle) \\
Wit_a(\epsilon) &\stackrel{\mathrm{def}}{=} (\epsilon = \langle a,-,-\rangle) \vee (\epsilon = \langle -,a,-\rangle)
\label{eq:respwit}
\end{split}
\end{equation}


\subsection{Semantics}
Based on possible world semantics, we define a frame
$\mathcal{F} = \langle\mathcal{W,B,D,I,E}\rangle$ as a tuple where:
\begin{itemize}
\item $W$ is a nonempty set of possible worlds,
\item $\mathcal{B}:AGT\rightarrow (W\rightarrow 2^W)$ is the function that associates each agent $a\in AGT$ and possible world $w\in W$ to the set of belief-accessible worlds $\mathcal{B}_a(w)$,
\item $\mathcal{D}:AGT\rightarrow (W\times DEG^+\rightarrow 2^W)$ is the function that associates each agent $a\in AGT$ and possible world $w\in W$ with a level of desirability $l\in DEG^+$ to the set of desire-accessible worlds $\mathcal{D}_a(w,l)$,
\item $\mathcal{I}:AGT\rightarrow (W\rightarrow 2^W)$ is the function that associates each agent $a\in AGT$ and possible world $w\in W$ to the set of intention-accessible worlds $\mathcal{I}_a(w)$, and
\item $\mathcal{E}:EVT\rightarrow W$ is the function that associates each event $\epsilon \in EVT$ to the resulting possible world.
\end{itemize}
Then, a model $\mathcal{M=\langle F,V\rangle}$ is a couple where $\mathcal{F}$ is a frame and $\mathcal{V}:W\rightarrow ATM$ a valuation function.  
\par Given a model $\mathcal{M}$ we note $\mathcal{M},w \models \varphi$ a formula $\varphi$ that is true in a world $w$. Truth conditions of formulas are defined by induction in the classical way:
\begin{itemize}
\item $\mathcal{M},w \models p$ iff $p\in \mathcal{V}(w)$;
\item $\mathcal{M},w \models \neg\varphi$ iff not $\mathcal{M},w \models \varphi$;
\item $\mathcal{M},w \models \varphi \wedge \psi$ iff $\mathcal{M},w \models \varphi$ and $\mathcal{M},w \models \psi$ ;
\item $\mathcal{M},w \models Bel_a^l(\varphi)$ iff $\frac{card(\mathcal{GB}_a(w))}{card(\mathcal{B}_a(w))}=l\;$ \\where $\mathcal{GB}_a(w) = \{v\in\mathcal{B}_a(w)$ ; $\mathcal{M},v \models \varphi \}$ ;
\item $\mathcal{M},w \models Des_a^l(\varphi)$ iff $\mathcal{M},v \models \varphi$ $\forall v\in\mathcal{D}_a(w,l)$;
\item $\mathcal{M},w \models Int_a(\varphi)$ iff $\mathcal{M},v \models \varphi$ $\forall v\in\mathcal{I}_a(w)$;
\item $\mathcal{M},w \models \epsilon$ iff $\mathcal{M},v \models \top$ $\forall v\in\mathcal{E}(\epsilon)$;
\end{itemize}
The truth condition of $Bel_a^l(\varphi)$ states that the level of plausibility of $\varphi$ is the proportion of belief-accessible worlds where $\varphi$ is true. The next subsections describe the rules for the inference engines presented in section \ref{sect:archi}. When required, the computation of believability, desirability and intensity degrees will be represented by a $f$ function that is part of the implementation and will not be detailed in this section (see section \ref{sect:implem} instead).


\subsection{Folk-psychology reasoner}
\subsubsection{Graded beliefs}
Following \cite{Dastani2012} and \cite{Adam2009}, all accessibility relations $\mathcal{B}$ are transitive and euclidean, which ensures that the agent is aware of its own beliefs\footnote{If $w\mathcal{R}v$ and $v\mathcal{R}u$, then successively by transitivity, euclidianity and transitivity again: $w\mathcal{R}v$ and $v\mathcal{R}v$.}:
\begin{equation}
Bel_a^l(\varphi) \stackrel{\mathrm{def}}{\Longrightarrow} Bel_a^1(Bel_a^l(\varphi))
\label{eq:beltreuclid}
\end{equation}
We generalize (\ref{eq:beltreuclid}) so that agents are aware of their own mental states, social relations and involvement.

However, unlike other models \cite{Adam2009} \cite{Dastani2012}, $\mathcal{B}$ is not serial\footnote{A relation $\mathcal{R}$ is serial iff $\forall w,\;\exists v$ so that $w\mathcal{R}v$. }. Only $\mathcal{GB}$ is. This represents the fact that the agent generally has uncertainty about states of affairs. Intuitively:
\begin{equation}
Bel_a^l(\varphi) \stackrel{\mathrm{def}}{\Longrightarrow} Bel_a^{1-l}(\neg\varphi)
\label{eq:belnotphi}
\end{equation}

For convenience, we define two thresholds $mod\_th$ and $str\_th$ wich $0.5 < mod\_th < str\_th$. They correspond to situations where the agent \textit{moderately} ($Bel_a^{l>mod\_th}(\varphi)$) and \textit{strongly} ($Bel_a^{l>str\_th}(\varphi)$) believes something.

Finally, if an agent believes a state of affairs to possibly cause another, it will deduce a belief about it:
\begin{equation}
Bel_a^l(\psi)\wedge Bel_a^{l'}(\psi \Rightarrow \varphi) \stackrel{\mathrm{def}}{\Longrightarrow} Bel_a^{f(l,l')}(\varphi)
\label{eq:newbel}
\end{equation}

\subsubsection{Graded attitudes}
Attitudes can be positive or negative and we assume that agents hold consistent desires:
\begin{equation}
\mathcal{M},w \models (Att_a^{k}(\varphi) \wedge Att_a^{k'}(\neg\varphi)) \,\mathrm{iff}\, k = -k'
\end{equation}

However, \textit{indirect} inconsistency is still possible: an agent might want something that can possibly lead to or be caused by (the occurrence of) the negation of another desire of his. The consistency is then preserved at the level of desire adoption:
\begin{equation}
\begin{split}
Des^k_a(\varphi) \wedge Bel^{l>str\_th}_a(\psi \Rightarrow F(\varphi)) \wedge \neg IncDes^k_a(\psi) \\ \stackrel{\mathrm{def}}{\Longrightarrow} N(Des^k_a(\psi))
\label{eq:newdes}
\end{split}
\end{equation}
with $IncDes^k_a(\varphi)$ representing inconsistent desires:
\begin{equation}
\begin{split}
IncDes^k_a(\varphi) \stackrel{\mathrm{def}}{=} 
&(Bel^{l>str\_th}_a(\varphi \Rightarrow \neg\psi) \wedge Des^{k'>0}_a(\psi)) \\ \vee &(Bel^{l>str\_th}_a(\varphi \Rightarrow \psi) \wedge Des^{k'<0}_a(\psi)) 
\label{eq:inconsdes}
\end{split}
\end{equation}

This means that desiring $\varphi$ is inconsistent when the agent
strongly beliefs it might lead to an undesirable $\psi$. We allow for
adopting indirectly inconsistent desires only when the agent only
believes \textit{moderately} that there can be a certain
incompatibility with existing ones.

We also define a weaker case of inconsistency where $\varphi$ leads to
an undesirable state of affairs of a higher level:
\begin{equation}
\begin{split}
WIncDes^k_a(\varphi) \stackrel{\mathrm{def}}{=} Bel^{l>str\_th}_a(\varphi \Rightarrow \neg\psi)
\\ \wedge Des^{k';|k'| > |k|}_a(\psi)
\end{split}
\label{eq:winconsdes}
\end{equation}

\subsubsection{Goals}
Following the BDI model \cite{rao1991modeling}, goals are defined as
desires that are consistent -- at least weakly, in our case -- and
believed to be achievable. To this purpose, we introduce a new
threshold $des\_th$:
\begin{equation}
Goal^{k>0}_a(\varphi) \stackrel{\mathrm{def}}{=} Des^{k>des\_th}_a(\varphi) \wedge Bel^{l}_a(F(\varphi)) \wedge \neg WIncDes^k_a(\varphi)
\label{eq:goal}
\end{equation}

Goals are then turned into intentions either because the agent can achieve it:
\begin{equation}
Goal^{k>0}_a(\epsilon) \wedge Resp_a(\epsilon) \stackrel{\mathrm{def}}{\Longrightarrow} N(Int_a(\epsilon))
\label{eq:goalintevt}
\end{equation}
or, similarly to \cite{bosse2011recursive}, because the agent strongly believes there is -- at least -- one mean to achieve it:
\begin{equation}
\begin{split}
Goal^{k>0}_a(\varphi) &\wedge Bel^{l>str\_th}_a(\psi \Rightarrow F(\varphi)) \wedge \neg WIncDes^k_a(\psi) 
\\ &\wedge Bel^{l'}_a(F(\psi)) \stackrel{\mathrm{def}}{\Longrightarrow} N(Int_a(\psi))
\label{eq:goalintphi}
\end{split}
\end{equation}
We leave it to the implementation phase (section~\ref{sect:implem}) to decide how intentions are ordered when several possible known $\psi$ can be used to achieve a goal.

\subsubsection{Intentions and acts}
Since intentions are generated from desires all accessibility relations $\mathcal{I}$ are serial: $\mathcal{M},w \not\models Int_a(\neg\varphi)$ if $\mathcal{M},w \models Int_a(\varphi)$

If an agent intends a state of affairs and knows a means to achieve it, it will also intend the latter:
\begin{equation}
Int_a(\varphi) \wedge Bel^{l>str\_th}_a(\psi \Rightarrow F(\varphi)) \stackrel{\mathrm{def}}{\Longrightarrow} Int_a(\psi)
\label{eq:newint}
\end{equation}
Additionally, if an agent intends an act which it is responsible for, it will perform it in the next step:
\begin{equation}
Int_a(\epsilon) \wedge Resp_a(\epsilon) \stackrel{\mathrm{def}}{\Longrightarrow} N(\epsilon)
\label{eq:intevt}
\end{equation}

Furthermore, when an event occurs, we propagate responsibility to all the states of affairs it is believed to have caused:
\begin{equation}
\begin{split}
Bel_a^d(\psi)\wedge Bel_a^l(Resp_b(\psi)) \wedge Bel_a^{l'}(\varphi) \wedge Bel_a^{l''}(\psi \Rightarrow F(\varphi)) 
\\\stackrel{\mathrm{def}}{\Longrightarrow} Bel_a^{f(l,l',l'')}(Resp_b(\varphi))
\label{eq:rspphi}
\end{split}
\end{equation}
\par Finally, as far as accessibility relations $\mathcal{E}$ are
concerned, we consider any witness believes with degree 1 that the
event happened and that the other witness also believes it. Note that
when an event occurs, the belief that it happened remains true
afterwards.

\subsubsection{Updating attitudes}
Beliefs are updated as new events occur (except for ideals that are
constant and hold globally). In order for the agent to react to
situation change, attitudes about new states of affairs have to be
triggered. In our model, following
\cite{ochs2009simulation,castelfranchi1998modelling,aylett2008if}, the
attitude is influenced not only by new beliefs, but also by the
attitude of others and the social relation. Formally:
\begin{equation}
\begin{split}
Bel^{l>str\_th}_a(\varphi) \wedge Att^k_a(F(\varphi)) \wedge Bel^{l'}_a(Att^{k'}_b(F(\varphi))) 
\\ \wedge Like^{h}_{a,b} \wedge Dom^{h'}_{a,b} \stackrel{\mathrm{def}}{\Longrightarrow} Att^{f(k,k',h,h')}_a(\varphi) \\
Bel_a^{l>str\_th}(Des_b^{k}(\varphi)) \wedge Like_{a,b}^{k'>0} \stackrel{\mathrm{def}}{\Longrightarrow} N(Des_a^{f(k,k')}(\varphi)))
\end{split}
\label{eq:attcurr}
\end{equation}

\subsubsection{Speech acts and social interaction}
Beliefs can also be updated through communication. Although our work
mostly focus on non-verbal communication, we consider a limited set of
illocutionnary acts \cite{searle1969speech} $ILL = \{Assert, Request,
Commit, Express\}$.

Based on similar work in speech acts formalization \cite{herzig2002logic,Guiraud2011}, we define trigering rules for our speech acts. For instance:
\begin{equation}
\neg Bel_a^1(Int_b(\varphi)) \wedge Int_a(Int_b(\varphi)) \stackrel{\mathrm{def}}{\Longrightarrow} Request_{a,b}(\varphi) 
\label{eq:spact}
\end{equation}


In turn, these events will lead to new mental states for the recipient agent, similarly to classical FIPA semantics and existing work on social interaction modeling \cite{herzig2002logic,castelfranchi1998modelling}. For the sake of conciseness, we only present these two examples here:
\begin{equation}
\begin{split}
Assert_{b,a}(\varphi) \wedge Like_{a,b}^{k}\wedge Dom_{a,b}^{k'} &\stackrel{\mathrm{def}}{\Longrightarrow} N(Bel_a^{f(k,k')}(\varphi))\\
Request_{b,a}(\varphi) \wedge Dom_{a,b}^{k<0} &\stackrel{\mathrm{def}}{\Longrightarrow} N(Int_a(\varphi)))
\end{split}
\label{eq:perloc}
\end{equation}




\subsection{Emotional inference engine}
The emotional inference engine consists of a set of appraisal rules
for emotion categories $EMO$. In this implementation, we have used an OCC-based model, highly inspired by
\cite{Adam2009,Guiraud2011,Dastani2012}. Here are some examples of triggering conditions for each group of emotions.

\begin{equation}
\begin{split}
Bel_a^l(\gamma) \wedge Att_a^{k>0}(\gamma) &\stackrel{\mathrm{def}}{\Longrightarrow} N(Joy_a^{i=f(l,k)}(\gamma)) \\
Bel_a^l(F(\gamma)) \wedge Des_a^{k<0}(\gamma) & \stackrel{\mathrm{def}}{\Longrightarrow} N(Fear_a^{i=f(l,k)}(\gamma))\\
Bel_a^d(\gamma)  \wedge Bel_a^l(Att_b^{k<0}&(\gamma)) \wedge Like_{a,b}^{k'<0} \\ & \stackrel{\mathrm{def}}{\Longrightarrow} N(Gloating_{a,b}^{i=f(l,k,k',d)}(\gamma)) \\
Bel_a^l(\gamma) \wedge Ideal_a^k(\gamma) \wedge &Bel_a^{l'}(Rsp_b(\gamma)) \\ & \stackrel{\mathrm{def}}{\Longrightarrow} N(Admiration_{a,b}^{i=f(l,l',k)}(\gamma))  \\
Bel_a^l(\gamma) \wedge Ideal_a^k(\gamma) \wedge &Bel_a^{l'}(Rsp_b(\gamma)) \wedge Goal_a^{k'}(\gamma)\\ & \stackrel{\mathrm{def}}{\Longrightarrow} N(Gratitude_{a,b}^{i=f(l,l',k, k')}(\gamma)) 
\label{eq:emo}
\end{split}
\end{equation}

Please note that $\gamma$ is a proposition that do not involve any temporal
operator. Besides, the intensity of an emotion is a
combination of the degree of certainty of beliefs and the degree of
desirability in attitudes. Depending on the appraisal model, the
degree of certainty can represent the \textit{sense of reality}, the
\textit{unexpectedness}, the \textit{likelihood} and the
\textit{realization}. The degree of desirability can correspond to
\textit{desirability-for-self} but also to \textit{praiseworthiness}
\cite{Adam2009}.

\par In OCC \cite{Ortony1990}, \textit{Gratification}, \textit{Remorse},
\textit{Gratitude} and \textit{Anger} are defined as
Well-being/Attribution emotions, triggered when one focuses both on
the praiseworthiness of an action and on its desirability. However, in
our model, these two notions overlap since ideals are deduced from
attitude. Nevertheless, similarly to \cite{Guiraud2011} we think that
one might distinguish \textit{Gratitude} and \textit{Anger} from
\textit{Admiration} and \textit{Reproach} if the triggering state of
affairs corresponds to a goal, that is to say it is not only
praiseworthy but is also desirable and consistent enough to generate
an intention of achievement.


\subsection{Commonsense reasoner}
The commonsence reasoner allows the agent to acquire new beliefs based
on a set of commonsense rules. It is mostly domain-dependent. Section
\ref{sec:job} describes how we used it to implement a job interview
simulation scenario. Here is a simple example of how this reasoner can
combine with the folk-psychology inference engines
presented above.
\subsubsection{Example} 
Consider two friends John (J) and Mary (M) having a conversation about
their holidays. Mary is going to her hometown (ht). The fact that she
is going to visit her father is a detail she could either mention or
not:
\begin{equation}
\nonumber
\begin{tabular*}{\linewidth}{l @{\extracolsep{\fill}} r}
$Des_{M}^{0.77}(talking\_about\_holidays)$ & (input)\\
$Bel_{M}^{0.8}(\langle M,J,visiting\_ht\_and\_dad \rangle$ \\$\hfill\Longrightarrow F(talk\_about\_holidays))$ & (input)\\
$Bel_{M}^{0.8}(\langle M,J,visiting\_ht \rangle $\\$\hfill\Longrightarrow F(talk\_about\_holidays))$ & (input)
\end{tabular*}
\end{equation}
Nevertheless Mary remembers John recently lost his father and thus supposes it is a sensitive topic:
\\\begin{equation}
\nonumber
\begin{tabular*}{\linewidth}{l @{\extracolsep{\fill}} r}
$Bel_{M}^{1}(J\_lost\_his\_dad)$ & (input)\\
$Bel_{M}^{0.76}(J\_lost\_his\_dad \Longrightarrow Ideal_{J}^{0.8}(\neg \langle -,J,dad \rangle) )$ & (input)\\
$\hspace{0.5cm} (\Longrightarrow) Bel_{M}^{l}(Ideal_{J}^{0.8}(\neg\langle -,J,dad \rangle))$ & (\ref{eq:newbel})
\end{tabular*}
\end{equation}
Of course, Mary knows that saying she is going to visit her father implies actually talking about her father:
\begin{equation}
\nonumber
\begin{tabular*}{\linewidth}{l @{\extracolsep{\fill}} r}
$Bel_{M}^{0.8}(\langle M,-,visiting\_ht\_and\_dad \rangle \hfill\Longrightarrow \langle M,-,dad \rangle )$ & (input)
\end{tabular*}
\end{equation}
And, knowing that John wants to avoid this topic, she does too. Hence, she is will not mention the fact that she is visiting her father when talking about her holidays:
\begin{equation}
\nonumber
\begin{tabular*}{\linewidth}{l @{\extracolsep{\fill}} r}
$\hspace{0.5cm} (\Longrightarrow) Ideal_{M}^{k}(\neg\langle -,J,dad \rangle)$ & (\ref{eq:newdes})\\
$\hspace{0.5cm} (\Longrightarrow)
WIncDes_{M}^{0.77}(\langle M,J,visiting\_ht\_and\_dad \rangle)$ & (\ref{eq:winconsdes}) \\
$\hspace{0.5cm} (\Longrightarrow)
Goal_{M}^{0.77}(\langle M,J,visiting\_ht \rangle)$ & (\ref{eq:goal}) 
\end{tabular*}
\end{equation}

\section{Implementation}
\label{sect:implem}

The theoretical model we presented in previous sections is aimed to be domain-independent. Yet, the purpose of our current work in the TARDIS project \cite{tardis2013} is to develop a training game in order to facilitate NEETs' access to employment. Therefore, we propose to implement it in the context of job interview simulation. Indeed, this sort of application appears as a promising way to increase applicant's self-confidence \cite{hoque2013}. Additionally, job interviews are a good example of semi-structured dyadic interactions where recruiters have several opportunities to reason about candidates' mental and affective states. Nevertheless, to our knowledge, existing models do not include a Theory of Mind. 

Our implementation was done in SWI-Prolog for the inference engine and
the logical framework. This reasoner was embedded in a C++ program
that handles the reasoning loop and the communication between the
modules.


\subsubsection*{Reasoning loop}
Following the classical BDI interpreter, at every cycle, the agent
interpret external events to generate a list of potential actions,
deliberate to select one of them, update its intentions and then
execute them:
\begin{algorithm}
\caption{ToM Reasoning loop}
\label{algo:loop}
\begin{algorithmic}
\small{
\Loop
	\State Execute\_intentions()
	\State Simulate\_others\_emotions()
	\State Update\_beliefs\_and\_attitudes()
	\State \hspace{\algorithmicindent} Update\_beliefs\_with\_new\_SoA()
	\State \hspace{\algorithmicindent} Handle\_operators\_equivalence()
	\State \hspace{\algorithmicindent} Adopt\_new\_desires()
	\State \hspace{\algorithmicindent} Order\_goals()
	\State Adopt\_new\_intentions()
	\State \hspace{\algorithmicindent} Adopt\_new\_intentions\_from\_goals()
	\State \hspace{\algorithmicindent} Adopt\_new\_intentions\_from\_intentions()
\EndLoop
}
\end{algorithmic}
\end{algorithm}

\subsubsection*{Thresholds and level functions}
The implementation of the model requires to instanciate all thresholds
($th$) and combination function ($f$) for degrees of believability and
desirability of new mental state or the intensity of emotions.

In our implementation, $mod\_th = 0.5$, $str\_th = 0.75$ and $des\_th
= 0.7$.

The combination function cannot be given in detail in this
paper but we consider two families:

-- For attitude dynamics and credibility (\textit{e.g.} equation
  \ref{eq:attcurr}), we use simple average functions on the relevant
  interval:
\begin{equation}
\nonumber
f(k,k') = ((k + k')/4)+0.5
\end{equation}

-- For emotions (see equation \ref{eq:emo}), we combine the linear
  influence from attitude (for instance, joy has been chosen to be
  linearly correlated to the attitude toward the fact) with a
  logarithmic influence of the degree of certainty. This way, we get
  to trigger more salient emotions even with relatively weak
  beliefs. Nevertheless, let us remind here that we only consider
  beliefs which levels are greater than a certain threshold
  ($mod\_thld=0.5$ in our implementation).

\begin{equation}
\nonumber
f(l,k) = \frac{k}{2} \times \frac{Log (2l-1) - min}{min}+0.5
\end{equation}
where $min$ represents the smallest value coded by the machine
(\textit{i.e.} the value of $Log(x)$ when $x\rightarrow 0$). The
$2l-1$ facto is used to adjust the value in [0,1] before we compute
the intensity, which is then readjusted in [0.5,1] to get significant
values.

\subsubsection*{Job interview simulation}
\label{sec:job}
The course of the job interview is handled in the commonsense module:
we define a series of topics that must be adressed by the agent
through speech acts (\textit{e.g.} questions about the salary, the
experience...). Moreover, each topic is associated with some
expectations about the impact of the question. Based on the current
\textit{goals} (in terms of affective state for the interlocutor), the
agent will select a question (a speech act) or another.

Moreover, the agent computes beliefs about the interlocutor's
self-confidence, motivation and qualification, based on its reaction
to the questions and simple TT-rules. For instance, hesitating in the
job description topic can indicate they are not qualified enough while
being focused when introducing themselves denotes a good
self-confidence level. The perception of ``hesitation'' and
``focused'' is done by another module of the TARDIS platform which is not part of this paper (see \cite{tardis2013}).


\section{Preliminary evaluation}
\label{sect:eval}
In this section, we describe a preliminary evaluation aiming to assess
the functioning of our model and its possible contribution in the
context of job interview simulation.


Subjects play the role of an unemployed youngster lacking work
experience and applying for the job of sales department secretary. The
virtual recruiter utterances are predefined for each possible speech act and situation in the model. No
constraint were given about a supposed personality, level of education
of professional background of the role-played interviewee.


\subsection{Method}
We recruited 30 volunteers -- 11 females and 19 males --, 19 of them working or having an internship at our university. All the subjects were aged over 24, had gone to university and were familiar with computers. 18 of the participants are native speakers and the remaining have at least an intermediate level. Since we are not interested in verbal communication, this is sufficient so that the participants understand what the recruiter says.

The recruiter's utterances were given in a very simple Graphical User Interface (GUI). The valence of the agent's affective state and its runtime evaluation of the candidate's self-confidence, motivation and qualification (values in [-1,1]) were represented by slide-bars. A text field allows the subject to type his/her answer to the virtual recruiter's questions. Besides, a series of 8 sliders (values in [0,1]) gives them the possibility to express their affective states to the recruiter as combinations of the following affects: relieved (REL), embarrassed (EMB), hesitating (HES), stressed (STR), ill at ease (IAE), focused (FOC), aggressive (AGG) and bored (BOR).\footnote{In the full TARDIS project's setting, these sliders are replaced by automatic recognition of user affects using the SSI system \cite{wagner2013social}.}


Subjects faced one agent out of 3 possible recruiter profiles: one that tries to make the candidates feel at ease (PROFILE\_A), one asking regular questions, with no specific goal on the user's mental state (PROFILE\_B) and one that, asks embarassing questions (PROFILE\_C). This is simply done by varying their goals regarding the emotional reaction they want to elicit in our model. All three agents use the same ToM reasoner described in previous sections. 

\par \textbf{Hypothesis:} The profile variation will have an impact on the participants' emotional states as expressed through the slidebars.
\par \textbf{Measures:} In this paper, we focus on measures extracted from the interaction history. They refer to the average intensity of relief, embarrassment, hesitation, stress, uneasiness, concentration, aggressiveness and boredom expressed by the participants as well as the total emotional expressiveness (TOT). More specifically, we measure the mean amount of information the candidates gave about their affective states.

\subsection{Results}
Shapiro-Wilks test shows that none of our measures follows a normal distribution. Besides, Kruskal-Wallis test reveals a main effect of PROFILE on TOT ($Chi^2(2,629)=11.435; p<0.01$) and particularly EMB ($Chi^2(2,629)=6.231; p<0.05$) and FOC ($Chi^2(2,629)=9.218; p<0.01$). This means that the profile of the recruiter (comprehensive, neutral or challenging) has an effect on the affects assessed (and possibly expressed) by the user, and that this effect is particularly important for embarassement and concentration.

\begin{figure}[h]
\centering
\includegraphics[width=\linewidth]{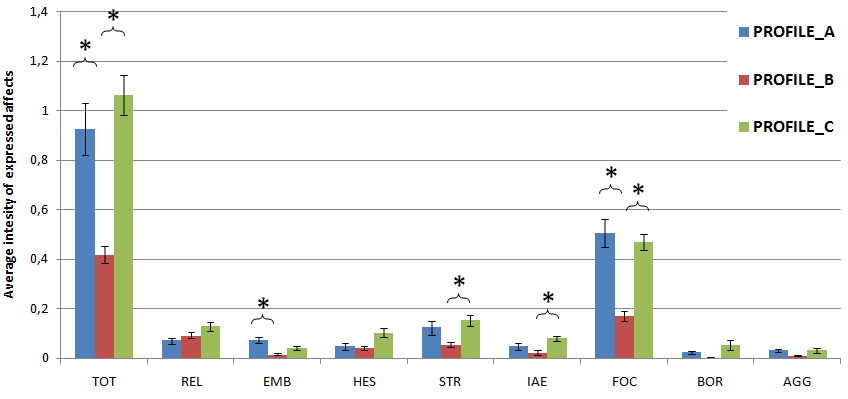}
\caption{Effect of PROFILE factor. This figure shows the average affects intensity expressed by the participants. The error bars represent the standard error. Significant effect appears on the embarrassment, stress, uneasiness, concentration and the total emotional intensity.}
\label{fig:resH2}
\end{figure}

A Mann-Whitney test then shows that participants that interact with PROFILE\_A (comprehensive recruiter) express more affects in general ($U = 20; p<0.05$), more embarrassment ($U = 20; p<0.05$) and more concentration ($U = 21; p<0.05$) than those who interact with PROFILE\_B. Likewise, PROFILE\_C (challenging recruiter) elicits more affects ($U = 6; p<0.01$) and in particular stress ($U = 18; p<0.05$), uneasiness ($U = 24; p<0.05$) and concentration ($U = 10; p<0.01$) than PROFILE\_B. We also note that in this case, no effect appears regarding embarrassment ($U = 26; p=0.069$). Finally, no significant effect is revealed between PROFILE\_A and PROFILE\_C. See Figure \ref{fig:resH2}.

\subsection{Discussion}
\label{sect:discuss}





Theory of Mind is a complex process that relies on various other cognitive and perceptual processes. It is not only hard to model but also to assess. Thus, a simple protocol such as the one we used in this study is not sufficient to fully evaluate the impact of our ToM model on the quality of the training. First, the GUI we used is not user-friendly and does not allow for user's immersion in the scenario. We assume that using the full TARDIS project's setting \cite{tardis2013} would enhance the interaction credibility and help highlight the virtual agent's reasoning and reactivity. Besides, the evaluation process should be based on richer measures (e.g. thorough post-hoc questionnaire) in order to evaluate the effect of our model. Yet, such a specific evaluation protocol for affective and interaction-oriented ToM still has to be defined.

In the litterature, there are validated methods to evaluate whether subjects -- generally children -- have ToM abilities and use it \cite{blijd2008measuring}. Nevertheless, there is no such test that integrates a strong interactional aspect to our knowledge. From the computational point of view, \cite{harbers2011explaining} points out the issue of evaluating a ToM model. In this work, the course of events and the agent's actions and explanations are specified in advance for different scenarios. Thus, the ToM models are evaluated based on whether they match these specifications. Similarly, \cite{pynadath2013you} builds expectations about user's actions -- based on formal models in the specific context of wartime negotiations -- in order to model a simplified theory of mind and then compare them with the actual user's behavior. These two approaches are not applicable in our human/agent interaction situation, because they rely on a model of the task which is difficult to describe when it comes to afective non-verbal behaviour. 

Nevertheless, the study we present in this paper shows promising results regarding the contribution of our ToM model in the context of a training game. Although all recruiters' profiles benefit from the ToM reasoner, only PROFILE\_A and PROFILE\_C use it to select questions according to a reasoning about the mental and emotional states they could induce. The more recruters ask such questions, the more mindreading they perform. The study shows that this kind of ToM-based behavior indeed has an impact on the users' reactions. It also demonstrates the benefit of implementing several profiles in the enrichment of the coaching scenarios.



\section{Conclusion \& perspectives}
\label{sect:conc}
In this paper, we proposed an affective and interaction-oriented Theory of Mind model to support the development of intelligent agents that are able to represent and reason about their human interlocutors' mental states. It relies on a hybrid mindreading approach mixing theory-theory and simulation-theory paradigms. This model is domain-independent, which means it can potentially be used in different context of application, including social coaching.

It has been implemented and evaluated in the context of job interview simulation in which the virtual recruiter both evaluates the human candidates based on their affective reactions, and reacts emotionally according to its desires and ideals. This study demonstrates the influence of the implementation of various recruiter profiles on the enrichment of the system's efficiency. In addition, the explanatory capability of our reasoning model is a key feature for the users to benefit from a rich post-interview feedback. While evaluating such a complex cognitive process as ToM remains a difficult task, we are currently working on the integration of this model in the TARDIS platform in order to perform mental states evaluation using signal processing. We plan to evaluate the impact of such a ToM model on the credibility of the virtual recruiter.
\bibliographystyle{abbrv}
\bibliography{collection}

\end{document}